\documentclass[a4paper,11pt]{extarticle}

\usepackage[utf8]{inputenc}

\usepackage[dvipsnames]{xcolor}

\usepackage{amsmath,amssymb,mathrsfs}
\usepackage[margin=1in]{geometry}
\usepackage{graphicx}
\usepackage{multirow}
\usepackage[export]{adjustbox}
\usepackage{bm}
\usepackage{float}
\usepackage{subcaption}
\usepackage{algorithm}
\usepackage[noend]{algpseudocode}
\usepackage{comment}
\usepackage{xcolor}
\usepackage{arydshln}
\usepackage{tikz}
\usepackage{pgfplots}
\usepgfplotslibrary{groupplots}
\usetikzlibrary{calc}
\usepgfplotslibrary{colorbrewer}
\usepgfplotslibrary{groupplots}

\pgfplotsset{compat=1.17}

\usetikzlibrary{pgfplots.groupplots}
\usetikzlibrary{matrix,arrows,decorations.pathmorphing}
\usepgfplotslibrary{fillbetween}

\long\def\comment#1{}

\usepackage{rays_defs_17}

\usepackage{hyperref}
\hypersetup{
    unicode=false,          
    pdftoolbar=true,        
    pdfmenubar=true,        
    pdffitwindow=false,     
    pdfstartview={FitH},    
    pdfauthor={Youssef Mansour},     
    pdfsubject={Subject},   
    pdfcreator={Youssef Mansour},   
    pdfproducer={Producer}, 
    pdfnewwindow=true,      
    colorlinks=true,       
    linkcolor=red,          
    citecolor=green,        
    filecolor=magenta,      
    urlcolor=cyan           
}

\usepackage[
backend=bibtex,
url=false,isbn=false,doi=false,
hyperref=auto,
style=alphabetic,
natbib=true,
sorting=nty,
firstinits=true,
maxnames=2, maxbibnames=10,
]{biblatex}
\bibliography{./references.bib} 

\title{\textbf{GAMA-IR: Global Additive Multidimensional Averaging for Fast Image Restoration}}
\author{Youssef Mansour and Reinhard Heckel \\
School of Computation, Information and Technology, Technical University of Munich\\
Munich Center for Machine Learning\\
 }
\date{}
\begin{document}
\maketitle

\begin{abstract}
Deep learning-based methods have shown remarkable success for various image restoration tasks such as denoising and deblurring. The current state-of-the-art networks are relatively deep and utilize (variants of) self attention mechanisms. Those networks are significantly slower than shallow convolutional networks, which however perform worse. 
In this paper, we introduce an image restoration network that is both fast and yields excellent image quality.
The network is designed to minimize the latency and memory consumption when executed on a standard GPU, while maintaining state-of-the-art performance. 
The network is a simple shallow network with an efficient block  that implements global additive multidimensional averaging operations. This block can capture global information and enable a large receptive field even when used in shallow networks with minimal computational overhead. 
Through extensive experiments and evaluations on diverse tasks, we demonstrate that our network achieves comparable or even superior results to existing state-of-the-art image restoration networks with less latency. For instance, we exceed the state-of-the-art result on real-world SIDD denoising by 0.11dB, while being 2 to 10 times faster.  
\end{abstract}

\maketitle

\section{Introduction}

\begin{figure}[t]
    \centering
    \begin{tikzpicture}
        \begin{groupplot}[
            group style={
                group size=2 by 1,
                horizontal sep=1.5cm,
            },
            width=0.52\textwidth,
            height=0.35\textwidth,
            xlabel style={yshift=-0.5em},
            ylabel near ticks,
            axis lines=left,
            grid=major,
            grid style=dashed,
            title style={yshift=-1.5ex},
        ]

        \nextgroupplot[
            ylabel={PSNR/dB},
            ylabel style={yshift=-0.5em},
            xmin = 10, xmax=150, ymin=39.6, ymax=40.5,
            xlabel = {Latency/ms}, 
            xlabel style={yshift=0.5em},
            xmode = log,
            title = {SIDD - Denoising}
        ]
        \addplot+[only marks, mark=star, mark size=3pt, mark options={blue},] coordinates {
        (15.1,40.02) 
        (20.7, 40.41) }; 
        \node[above, font=\small, text=blue] at (15.1,40.15) {GAMA-IR};

        \addplot+[only marks, mark=*, mark size=2pt, mark options={red},] coordinates {
        (30.1, 39.96) 
        (35.2, 40.30) }; 
        \node[above, font=\small, text=red] at (31.2,40.05) {NAFNet};

        \addplot+[only marks, mark=*, mark size=2pt, mark options={black},] coordinates {
        (92.1, 40.02) }; 
        \node[above, font=\small, text=black] at (92.1,40.02) {Restormer};       

         \addplot+[only marks, mark=*, mark size=2pt, mark options={brown},] coordinates {
        (133.7, 39.96) }; 
        \node[below, font=\small, text=brown] at (110, 39.96) {MAXIM};

         \addplot+[only marks, mark=diamond*, mark size=2pt, mark options={purple},] coordinates {
        (34.6, 39.89) }; 
        \node[below, font=\small, text=purple] at (34.6, 39.89) {Uformer};

         \addplot+[only marks, mark=diamond*, mark size=2pt, mark options={cyan},] coordinates {
        (27, 39.99) }; 
        \node[left, font=\small, text=cyan] at (27, 39.99) {HINet};


          \addplot+[only marks, mark=square*, mark size=2pt, mark options={green},] coordinates {
        (35.6, 39.71) }; 
        \node[below, font=\small, text=green] at (35.6, 39.71) {MPRNet};

        \nextgroupplot[
            ylabel={PSNR/dB},
            ylabel style={yshift=-0.5em},
            xmin = 10, xmax=150, ymin=32.1, ymax=33.2,
            xlabel = {Latency/ms}, 
            xlabel style={yshift=0.5em},
            xmode = log,
            title = {GoPro - Deblurring}
        ]
        \addplot+[only marks, mark=star, mark size=3pt, mark options={blue},] coordinates {
        (15.1,32.44) 
        (20.7, 33.15) }; 
        \node[below, font=\small, text=blue] at (15.8,33.15) {GAMA-IR};
        \node[above, font=\small, text=blue] at (14.7,32.44) {GAMA-IR};

        \addplot+[only marks, mark=*, mark size=2pt, mark options={red},] coordinates {
        (30.1, 32.29) 
        (35.2, 33.08) }; 
        \node[right, font=\small, text=red] at (30.01,32.29) {NAFNet};
        \node[right, font=\small, text=red] at (35.2,33.08) {NAFNet};

        \addplot+[only marks, mark=*, mark size=2pt, mark options={black},] coordinates {
        (92.1, 32.92) }; 
        \node[above, font=\small, text=black] at (92.1,32.92) {Restormer}; 

         \addplot+[only marks, mark=*, mark size=2pt, mark options={brown},] coordinates {
        (133.7, 32.86) }; 
        \node[below, font=\small, text=brown] at (110, 32.86) {MAXIM};     

         \addplot+[only marks, mark=diamond*, mark size=2pt, mark options={purple},] coordinates {
        (34.6, 32.97) }; 
        \node[below, font=\small, text=purple] at (34.6, 32.97) {Uformer};
        
         \addplot+[only marks, mark=diamond*, mark size=2pt, mark options={cyan},] coordinates {
        (27, 32.77) }; 
        \node[below, font=\small, text=cyan] at (27, 32.77) {HINet};

          \addplot+[only marks, mark=square*, mark size=2pt, mark options={green},] coordinates {
        (35.6, 32.66) }; 
        \node[below, font=\small, text=green] at (35.6, 32.66) {MPRNet};    

          \addplot+[only marks, mark=diamond*, mark size=2pt, mark options={magenta},] coordinates {
        (17, 32.45) }; 
        \node[below, font=\small, text=magenta] at (20, 32.45) {MIMO};  
        \node[below, font=\small, text=magenta] at (20, 32.35) {UNet+};  
        \end{groupplot}
    \end{tikzpicture}

\caption{Performance and speed of our GAMA-IR network in comparison to other popular and state-of-the-art networks on two common restoration tasks: Image denoising and deblurring. GAMA-IR achieves slightly better performance than state of the art networks, while being significantly faster. 
}
    \label{fig:intro}

\end{figure}
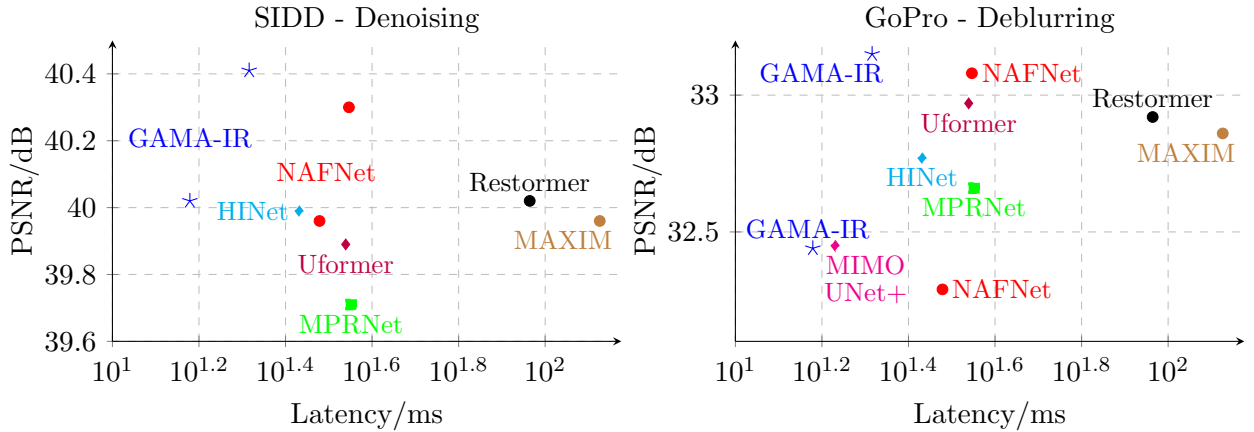

Deep learning gives state-of-the-art performance for a variety of image restoration tasks like denoising, deblurring, and deraining. 
Image restoration is often formulated as an image-to-image regression problem, i.e., to map a noisy image or a coarse reconstruction of an image to a clean image \cite{dncnn, brooks}. 

The network architectures that map the noisy or coarse reconstruction to the clean images play a pivotal role for performance. For many years, the prominent UNet \cite{unet} has been the go-to network for different reconstruction tasks, since it performs well and has low computational requirements. 
The UNet is therefore a common choice for image reconstruction problems where speed is a concern, and it is also frequently used as a building block in more complicated reconstruction approaches. For example, recent self-supervised denoising methods \cite{noise2noise,noise2void,noisier2noise,nb2nb,blind2unblind} as well as the Variational Network \cite{mri3}, a state-of-the art network for accelerated MRI, all utilize the UNet as a building block. 

In the recent years, several deep networks that use a self attention \cite{attention_all_you_need} mechanism or variants of it have been proposed for image reconstruction, in particular SwinIR \cite{swin_ir}, Restormer \cite{restormer}, and MAXIM \cite{maxim}. Those networks provide better reconstructions than the UNet at the cost  of significantly higher compute, due to their depth, and the computational cost of the attention or attention-like mechanisms used in those networks. 

There is large demand for efficient and simple network architectures that enable fast inference times, consume little memory, while maintaining state-of-the-art image restoration performance. We focus on latency and memory at inference instead of MACs (multiply-accumulate operations) or number of parameters, since in many practical applications, speed and memory consumption are the metrics of interest.


In this paper, we propose a network with state-of-the-art performance but with low latency and memory consumption. 
The network is shallow and only uses convolutions, non-linearities, skip connections \cite{resnet}, and LayerNorm \cite{layernorm}. A key component of the network is a simple global additive multidimensional averaging (GAMA) block that efficiently captures non-local information and enables a large receptive field even when used in shallow networks. 
We integrate this block into a relatively standard convolutional neural network architecture, and the resulting network is called GAMA-IR (IR for image restoration). GAMA-IR can compete with state-of-the-art models at higher speed as seen in Figure \ref{fig:intro}.

State-of-the-art networks like Restormer\cite{restormer} and NAFNet\cite{nafnet} are usually deep. The network's depth enables good performance, but causes the network to be slow, since computations must be performed sequentially across multiple layers. Therefore, the challenge in designing a fast network lies in enabling a shallow network to perform well, which our proposed GAMA block achieves.


Up to our knowledge, the only recent image restoration network to focus on minimizing the latency instead of the MACs is MalleNet \cite{malleable_conv}. MalleNet is a lightweight network that utilizes an efficient variant of dynamic convolutions \cite{dynam_conv} to create a fast network. Although MalleNet reaches impressive speed, its performance is notably below the state of the art. 

Recently, several other efficient networks have been proposed that focus on minimizing the number of parameters or FLOPs (floating point operations). However, the number of parameters and FLOPs are not necessarily well correlated with a network's latency \cite{repvgg, efficiency1, mobileone} and/or with a network's memory footprint as shown in Figure \ref{fig:correlation}. That is because metrics like parameter count and FLOPs do not consider the memory access cost and degree of parallelism, which can significantly affect a network's speed and memory expenditure \cite{efficiency2}. 

For instance, sharing a model's weights decreases the model's number of parameters, but does not change the latency. Moreover, adding skip connections \cite{resnet} does not change the parameter count and has a negligible effect on the FLOPs, but requires significantly more memory. This weak correlation between the metrics has been manifested by SwinIR \cite{swin_ir}, which is very parameter-efficient. However, the network is slow and consumes a lot of memory in comparison with other baseline networks as seen in Figure \ref{fig:correlation}. 

Contrary to the previous restoration works, our work is the first to directly target the compute metrics of interest on a GPU: Memory and speed. The main contribution of our work, illustrated in Figure \ref{fig:intro}, is: 
\begin{itemize}
    \item We optimise the network's latency and memory expenditure instead of the number of trainable parameters or FLOPs as in previous restoration works.      
    \item We propose a global additive multidimensional averaging block that efficiently captures global dependencies, and enables a shallow network to have a large receptive field.
    \item We design a simple yet effective image restoration network, that performs similar to the state of the art models, but at much higher speed. 
\end{itemize}




\section{Related Work}

\subsection*{Image-to-image Networks}

Convolutional neural networks are a powerful choice for image reconstruction. 
A simple stack of convolutional layers, known as the the DnCNN \cite{dncnn} architecture, works well for denoising. 
Today, the encoder-decoder based UNet \cite{unet} is a popular choice for its good performance-speed trade off. The UNet was initially proposed for medical image segmentation, but became popular for image reconstruction tasks, since it is faster and performs better than DnCNN~\cite{brooks}. Later, several variants were proposed such as Attention-UNet \cite{attention_unet} and UNet++ \cite{unet++}.

With the rise of the transformer architecture and the introduction of the vision transformer \cite{vit}, several networks started incorporating attention mechanisms, which improve imaging performance at a price of computational cost. 
The vision transformer itself performs well for image reconstruction, better than the UNet, but at a significantly higher computational cost \cite{vit_recon}. 
The transformer based SwinIR \cite{swin_ir}, that builds on the Swin Transformer \cite{swin}, exhibits excellent performance without any convolutions. Later, Uformer \cite{uformer} and Restormer \cite{restormer} showed even better performance when utilizing convolutions with attention based architectures. 

Inspired by the MLP-Mixer \cite{mlp_mixer}, which only uses MLPs for image classification, several MLP-based reconstruction networks were proposed. The Image-to-Image Mixer \cite{img2img_mixer} stacks several MLP layers without any convolutions or attention gates. MixerGAN \cite{mixergan} is an MLP based architecture for Unpaired Image-to-Image Translation. MAXIM \cite{maxim} is a multi-axis spatially gated MLP-based network that achieves remarkable performance. It is no surprise that MLP based networks work well; plain MLPs \cite{plain_mlp} can compete with BM3D \cite{bm3d}, a classical denoising algorithm.

\subsection*{Self Attention-based architectures}


Transformers and attention modules exhibited success in vision tasks \cite{attention_all_you_need, vit}, but self attention mechanisms exhibit quadratic complexity with the input size, which raises the cost significantly for high resolution inputs. Non-local blocks \cite{nonlocal_blocks} compute the response at a position as a weighted sum of the features at all positions, which performs well, but is expensive to execute. To reduce the cost, Squeeze-and-Excite blocks \cite{squeeze_excite} capture channel attention only by performing global average pooling across the channel axis. Efficient Channel Attention Net \cite{1D_channel_attention} simplifies Squeeze-and-Excite by removing the intermediate convolution layers, and utilizing 1D convolutions. Convolutional Block Attention Modules \cite{cbam} perform global max pooling in addition to average pooling across both channel and spatial axes. We devise a block that builds on previous works, and efficiently captures global information, and enables a shallow network to have a large receptive field.   

\subsection*{Efficient Networks in other Domains}

The efficiency of neural networks has been widely studied in other domains, such as classification and segmentation.
MobileNet \cite{mobilenet}, is a lightweight convolutional neural network that achieved impressive accuracy, while keeping the computational complexity low, even on mobile devices. Similarly, EfficientNet \cite{efficientnet} is designed to be a scalable and efficient network, and achieves good performance across various resource-constrained settings. Recently, MobileOne \cite{mobileone} was proposed, and was shown to outperform several high performing models in under one millisecond runtime.

Instead of designing lightweight networks, one can change the network's structure from train to test time, such that the network can benefit from large capacity during training, while being efficient at inference. For example, in knowledge distillation \cite{distillation}, a larger teacher network is used during training, and a smaller, often simpler, student network is used during inference. ExpandNets \cite{expandnets} are linearly over-parameterized networks with a larger teacher network guiding a smaller more efficient student network. RepVGG \cite{repvgg} combines multiple parallel branches during training into a single branch at inference, achieving better performance than the original VGG nets \cite{vgg} without extra compute at inference. 


\section{GPU Efficiency Metrics}

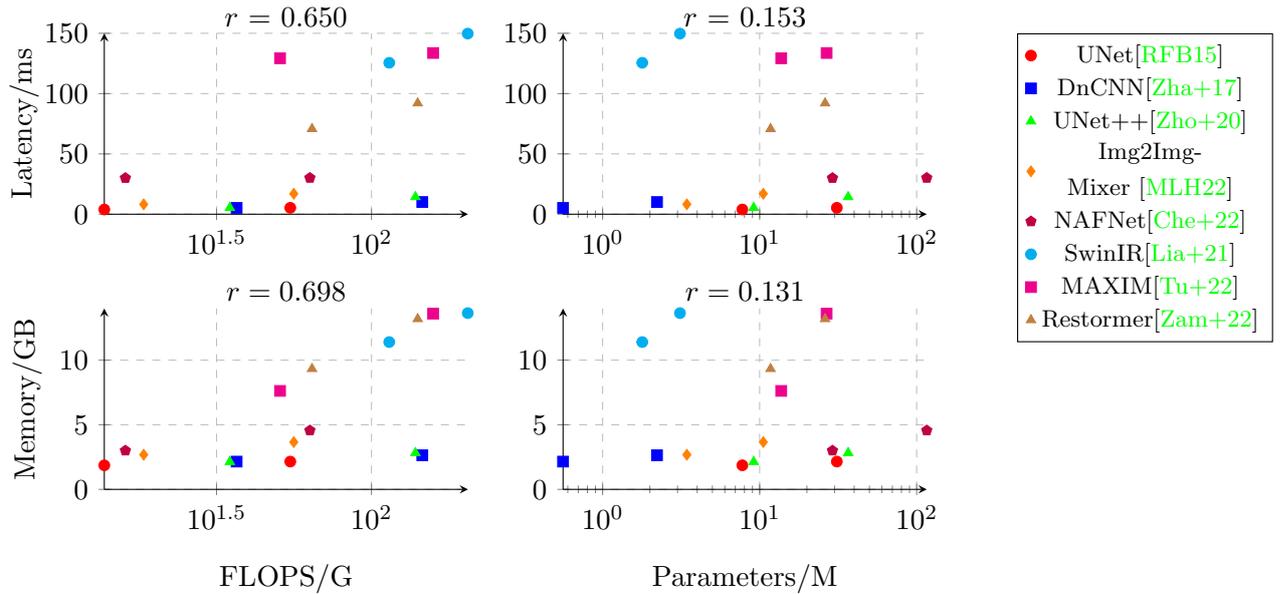
\begin{figure}[t]
    \centering
    \begin{tikzpicture}
        \begin{groupplot}[
            group style={
                group size=2 by 2,
                horizontal sep=1.25cm,
                vertical sep=1.25cm,
            },
            width=0.4\textwidth,
            height=0.25\textwidth,
            xlabel style={yshift=-0.5em},
            ylabel near ticks,
            axis lines=left,
            grid=major,
            grid style=dashed,
            title style={yshift=-1.5ex},
            legend style={at={(1.6,1)},anchor=north}, 
            legend style={cells={align=center}},
        ]

        \pgfplotscreateplotcyclelist{mylist}{
            {red, mark=*},
            {blue, mark=square*},
            {green, mark=triangle*},
            {orange, mark=diamond*},
            {purple, mark=pentagon*},
            {cyan, mark=oplus*},
            {magenta, mark=square*},
            {brown, mark=triangle*},
        }
        
        \nextgroupplot[
            ylabel={Latency/ms},
            ylabel style={yshift=-0.5em},
            xmax=205, ymin=0, ymax=150,
            cycle list name=mylist,
            xmode = log,
            title = {$r$ = 0.650}
        ]
        \addplot+[only marks, mark size=2pt] coordinates {
        (13.73,3.90)
        (54.70, 5.31)  }; 
        \addplot+[only marks, mark size=2pt] coordinates {
        (36.72, 5.21)
        (145.91, 10.10)};
        \addplot+[only marks, mark size=2pt] coordinates {
        (34.91, 5.35)
        (138.67, 14.30) };
        \addplot+[only marks, mark size=2pt] coordinates {
        (18.40, 8.19)
        (56.14, 16.95)};
        \addplot+[only marks, mark size=2pt] coordinates {
        (16.06, 30.01)
        (63.28, 30.11)};
        \addplot+[only marks, mark size=2pt] coordinates {
        (114.11, 125.55)
        (204.53, 149.74)};
        \addplot+[only marks, mark size=2pt] coordinates {
        (158.16, 133.66)
        (50.69, 129.28)};
        \addplot+[only marks, mark size=2pt] coordinates {
        (140.99, 92.05)
        (64.29, 70.74)
        };
        \nextgroupplot[xmax=116, ymin=0, ymax=150, cycle list name=mylist,xmode=log, title = {$r$ = 0.153}
        ]
        \addplot+[only marks, mark size=2pt] coordinates {
        (7.76,3.90)
        (31.03, 5.31)};
        \addplot+[only marks, mark size=2pt] coordinates {        
        (0.56, 5.21)
        (2.22, 10.10)};
        \addplot+[only marks, mark size=2pt] coordinates {
        (9.16, 5.35)
        (36.63, 14.30)};
        \addplot+[only marks, mark size=2pt] coordinates {
        (3.44, 8.19)
        (10.55, 16.95)};
        
        \addplot+[only marks, mark size=2pt] coordinates {
        (29.10, 30.01)
        (115.86, 30.11)};
        \addplot+[only marks, mark size=2pt] coordinates {
        (1.79, 125.55)
        (3.11, 149.74)};
        \addplot+[only marks, mark size=2pt] coordinates {
        (26.70, 133.66)
        (13.72, 129.28)};
        \addplot+[only marks, mark size=2pt] coordinates {
        (26.11, 92.05)
        (11.73, 70.74)
        };

        \addlegendentry{\footnotesize UNet\cite{unet}}
        \addlegendentry{\footnotesize DnCNN\cite{dncnn}}
        \addlegendentry{\footnotesize UNet++\cite{unet++}}
        \addlegendentry{\footnotesize Img2Img- \\ \footnotesize Mixer \cite{img2img_mixer}}
        \addlegendentry{\footnotesize NAFNet\cite{nafnet}}
        \addlegendentry{\footnotesize SwinIR\cite{swin_ir}}
        \addlegendentry{\footnotesize MAXIM\cite{maxim}}
        \addlegendentry{\footnotesize Restormer\cite{restormer}}

        \nextgroupplot[
            ylabel={Memory/GB},
            ylabel style={yshift=-0.1 em},
            xlabel={FLOPS/G},
            xmax=205, ymin=0, ymax=14,
            cycle list name=mylist,
            xmode=log,
            title = {$r=0.698$}
        ]
        \addplot+[only marks, mark=*, mark size=2pt] coordinates {
        (13.73,1.86)
        (54.70, 2.15)};
        \addplot+[only marks, mark size=2pt] coordinates {       
        (36.72, 2.15)
        (145.91, 2.64)};
        \addplot+[only marks, mark size=2pt] coordinates {
        (34.91, 2.13)
        (138.67, 2.82)};
        \addplot+[only marks, mark size=2pt] coordinates {
        (18.40, 2.68)
        (56.14, 3.66)};
        \addplot+[only marks, mark size=2pt] coordinates {
        (16.06, 3.00)
        (63.28, 4.56)};
        \addplot+[only marks, mark size=2pt] coordinates {
        (114.11, 11.40)
        (204.53, 13.64)};
        \addplot+[only marks, mark size=2pt] coordinates {
        (158.16, 13.60)
        (50.69, 7.62)};
        \addplot+[only marks, mark size=2pt] coordinates {
        (140.99, 13.19)
        (64.29, 9.33)
        };
        \nextgroupplot[
            xlabel={Parameters/M},
            xmax=116, ymin=0, ymax=14,
            cycle list name=mylist, xmode=log, title = {$r=0.131$}
        ]
        \addplot+[only marks, mark size=2pt] coordinates {   
        (7.76,1.86)
        (31.03, 2.15)};
        \addplot+[only marks, mark size=2pt] coordinates {           
        (0.56, 2.15)
        (2.22, 2.64)};
        \addplot+[only marks, mark size=2pt] coordinates {   
        (9.16, 2.13)
        (36.63, 2.82)};
        \addplot+[only marks, mark size=2pt] coordinates {   
        (3.44, 2.68)
        (10.55, 3.66)};
        \addplot+[only marks, mark size=2pt] coordinates {   
        (29.10, 3.00)
        (115.86, 4.56)};
        \addplot+[only marks, mark size=2pt] coordinates {   
        (1.79, 11.40)
        (3.11, 13.64)};
        \addplot+[only marks, mark size=2pt] coordinates {   
        (26.70, 13.60)
        (13.72, 7.62)};
        \addplot+[only marks, mark size=2pt] coordinates {   
        (26.11, 13.19)
        (11.73, 9.33)
        };

        \end{groupplot}
    \end{tikzpicture}
  
\caption{Correlation of FLOPs and parameter count with latency and memory of image restoration algorithms run on a NVIDIA384RTX A600 GPU. Latency and memory are measured as the time and memory required for a forward pass through the network at inference. All metrics are considered for an image input size of $1\times 3 \times 256 \times 256$, which is a single 3-channel RGB image (i.e., the batch size is one). The networks' parameters are varied to create different sizes of the same network. Note the moderate to weak correlation between FLOPS and Parameters with Memory and Latency.}

    \label{fig:correlation}
\end{figure}

We propose a network that reaches a good trade-off between image quality and computational efficiency, specifically we focus on latency and memory consumption on a GPU. 
Note that a network can be computationally efficient with respect to one metric but inefficient with respect to another~\cite{efficiency1}. 
Several works measure a network's efficiency with FLOPs/MACs or number of parameters. 
However, a network with a small FLOP count or with few parameters can still be significantly slower and consume significantly more memory when run on a GPU, relative to a network with more FLOPs.



For many applications, image restoration methods run on a GPU, and in such a setup, speed and memory consumption are the metrics of interest, since they 
provide a more accurate measure for the actual cost spent to deploy the network on a GPU. 

In Figure~\ref{fig:correlation} we plot the correlation between FLOPs and parameter count (commonly reported efficiency metrics) with speed and memory consumption on a GPU (metrics of interest when run on a GPU). We plot the FLOPs and number of parameters against the memory and latency for several image reconstruction networks ranging from classical models to newer state-of-the-art ones. We also calculate Spearman's rank correlation coefficient $r$ for every plot. By inspecting the plots and correlation factors, we can see that FLOPs are moderately correlated with memory and latency, whereas parameter count is only very weakly correlated.


\section{GAMA-Image Restoration (IR) Network}


In this section we introduce the GAMA-IR network. The key to achieving state-of-the-art performance at significantly higher speed than competing networks is to limit the depth of the network. This poses a challenge, since high performance is usually attained by deep networks which often have a high receptive field. 

Our GAMA block enables a large receptive field even when used in a shallow network, therefore enabling state-of-the-art performance at higher speed. We start with defining the overall architecture of our network, and then introduce the GAMA block.


\begin{figure*}[t]
\centering
{\includegraphics[width=\linewidth]{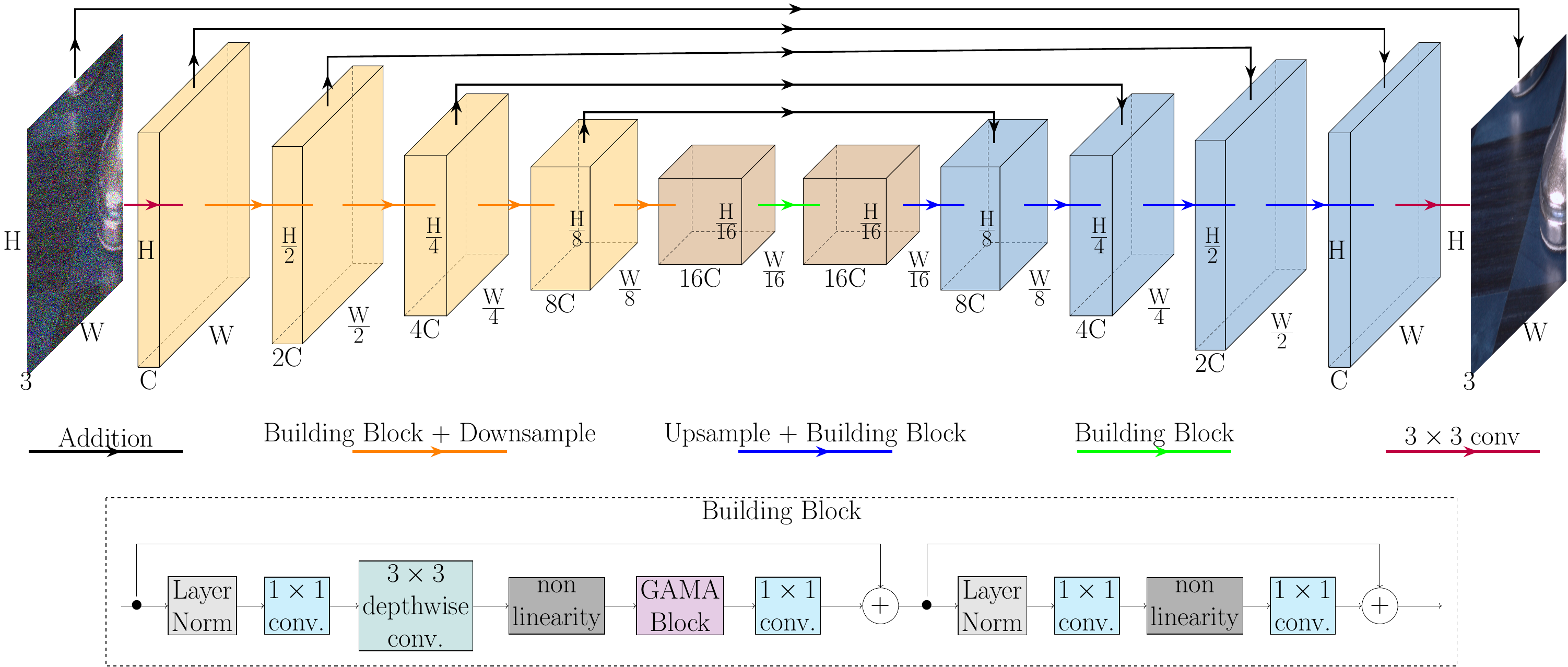}}
\caption{
\label{fig:network}
The architecture of our proposed network. Similar to the UNet, the network has an encoder-decoder structure with skip connections adding the feature maps of equal resolution elementwise. The multi-resolution layout is achieved by downsampling and upsampling operations. The network has only 2 hyperparameters, the depth and width. The depth is the total number of building blocks, and the width is the number of channels $C$. 
}
\end{figure*}

\begin{figure}[t]
\centering
{\includegraphics[width=\linewidth]{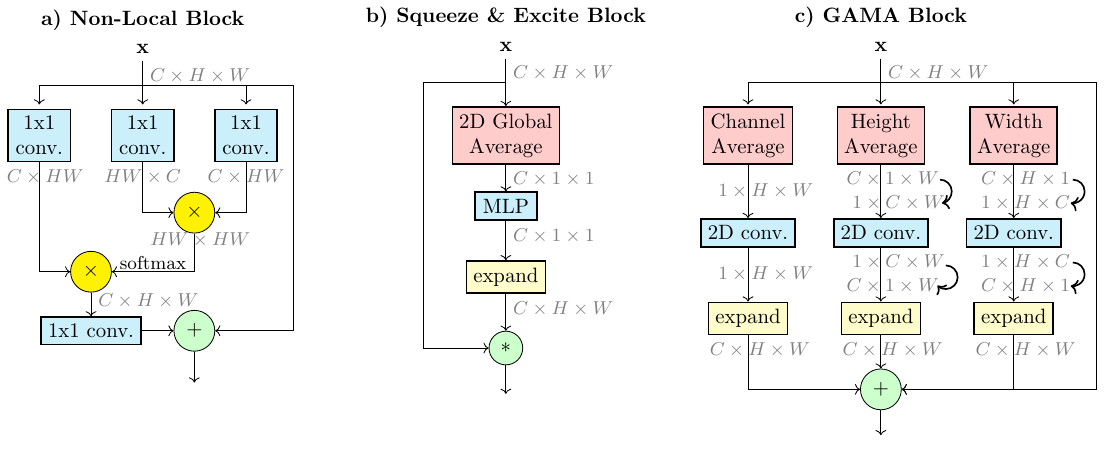}}
\caption{
\label{fig:blocks}
a) Non-Local Block \cite{nonlocal_blocks}. b) Squeeze-and-Excite Block \cite{squeeze_excite} c) Our proposed GAMA Block. $\otimes
$ denotes matrix multiplication, $\oplus$ and $\scalebox{0.75}{\textcircled{{\raisebox{-2.7pt}{*}}}}$ denote element wise addition and multiplication, respectively. The circular arrows in \textbf{c)} denote a transpose operation. The blocks aid in capturing long-range dependencies in data. Non-local blocks calculate interactions between all pixels (matrix multiplications), while the other blocks rely on global averaging to summarize the entire feature map across a specific dimension.
}
\end{figure}

GAMA-IR is is displayed in Figure \ref{fig:network}. The input image is first converted to a feature map with $C$ channels using a $3\times3$ convolution. The feature map is then processed by a series of building blocks and downsampling and upsampling layers. Skip connections from the encoder to the decoder add the feature maps of equal resolution element wise. The last feature map is mapped back to the image space via a final $3\times3$ convolution. 

We follow the down and upsampling scheme of Pixel Shuffle \cite{pixel_shuffle}, but use $1 \times 1$ convolutions instead of $3 \times 3$ to reduce cost. When downsampling, the number of channels are doubled and the height and the width are halved. This is achieved by a $1 \times 1$ convolutional layer that first maps a feature map of size $C \times H \times W$ to $ \frac{C}{2} \times H \times W$, which is then reshaped to $ 2C \times \frac{H}{2} \times \frac{W}{2}$. When upsampling, those computations are reversed, specifically, a feature map of size $C \times H \times W$ is mapped via $1 \times 1$ convolutions to $2C \times H \times W$, which is then reshaped to $\frac{C}{2} \times 2H \times 2W$.

The main operations are carried out by a building block that consists of our proposed GAMA block and standard convolutional network components, such as convolutional operations, non-linearities, skip connections \cite{resnet}, and LayerNorm \cite{layernorm}. 
 

Following Restormer \cite{restormer}, we use $1 \times 1$ convolutions followed by $3 \times 3$ depthwise convolutions \cite{depthwise_conv} in the building block. This combination of convolutions can be thought of as a sparse standard pointwise $3 \times 3$ convolution, that performs equally well, but with substantially less parameters and memory consumption. 

We adopt NAFNet's \cite{nafnet} non-linearity, which was shown to slightly outperform classical activation functions such as ReLU \cite{relu} or GeLU \cite{gelu}. The non-linearity splits the input feature map along the channel axis in two halves, and then multiplies the two halves element wise. 

The key difference to existing restoration networks is the GAMA block that models global dependencies. We first describe the global blocks of other state-of-the-art networks, and then explain the GAMA block, and finally provide a comparison for the receptive field of the different blocks. 

\subsection{Global Blocks}
\label{sec:global_blocks}

Convolutional networks operate locally only, and have a limited receptive field, except when concatenated several times which leads to deep networks. This limits their ability to capture interactions between far-away pixels, making them inherently local in their processing. To enable global dependencies, Restormer \cite{restormer} utilizes a variant of the Non-Local Block \cite{nonlocal_blocks}, illustrated in Figure \ref{fig:blocks}.

Non-local blocks are designed to capture long-range dependencies in data by utilizing self attention \cite{attention_all_you_need}. They perform matrix multiplications which allow each position of the input feature map to attend to 
every other position. This helps capture global context and dependencies in the data, making them useful for imaging tasks. However, the execution time of these matrix multiplications is high, making non-local blocks resource intensive and challenging to scale for large inputs.

Instead of calculating all pairwise interactions, NAFNet \cite{nafnet} utilizes a Squeeze-and-Excite block~\cite{squeeze_excite}, that executes 2D global averaging, which squeezes a 3D feature map of size $C \times H \times W$ to a 1D channel vector of length $C$, i.e., each spatial feature $H \times W$ is reduced to a scalar value. The channel vector is passed through an MLP (multi-layer perceptron) to output a new vector of same length, which is multiplied element-wise with the input feature map. This is equivalent to multiplying each channel of the input feature map by a learnable scalar weight, i.e., giving more weight to more relevant channels, while suppressing less informative ones. In summary, a Squeeze-and-Excite block captures global dependencies across the channel axis.


A Squeeze-and-Excite block is significantly faster than a Non-local block, since it does not perform matrix multiplications, but only global averaging and element-wise multiplication. However, a major drawback is that it only captures channel-wise dependencies, which significantly increases the number of parameters. The number of parameters in the MLP is $C^2$. Since in deep layers of a network, $C$ exceeds $1000$, this significantly increases the parameters per block. Our proposed GAMA block aids in overcoming the aforementioned drawbacks. 

\subsection{GAMA Block}
\label{sec:GAMA_Block}

The input to the GAMA block, a feature map of size $C \times H \times W$, is globally averaged across each dimension separately, resulting in three feature maps of sizes $1 \times H \times W$, $C \times 1 \times W$, and $C \times H \times 1$. The latter two are transposed to $1 \times C \times W$ and $1 \times H \times C$ respectively. The three feature maps are convolved with three separate 2D convolutions with input and output channels = 1, and kernel size = 7. Finally, the resulting feature maps are transposed and then expanded to match the initial input size ($C \times H \times W$) before being added together with the input feature map to generate the block's output as seen in figure~\ref{fig:blocks}. The expansion occurs by simply replicating the values across the appropriate dimensions.

Unlike Squeeze-and-Excite, we capture global information from the entire feature map, since we average along all three dimensions. We also execute global averaging across one dimension at a time, which results in 2D feature maps instead of one dimensional vectors, and therefore less loss of information. Moreover, our block has minimal parameter overhead.
Each 2D convolution operator has $7 \times 7 \times 1 \times 1 = 49$ learnable parameters, which results in only $3 \times 49 = 147$ parameters per block. The number of parameters is fixed, and independent of the size of the input feature map.

\subsection{Receptive Field}

\newcommand\wwww{0.195}
\begin{figure*}[t]
  \captionsetup[subfigure]{labelformat=empty,justification=centering}
  \centering

  \includegraphics[scale=0.4]{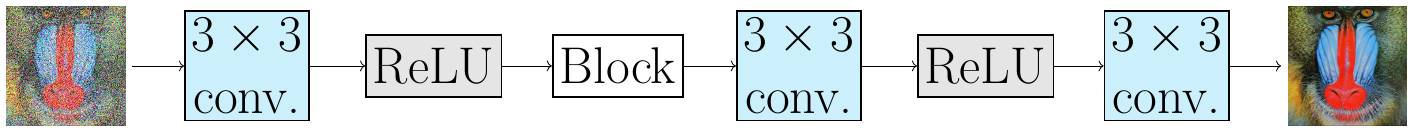}

  \makebox[\wwww\linewidth]{ Clean Image} \hfill
  \makebox[\wwww\linewidth]{\footnotesize Noisy Image} \hfill
  \makebox[\wwww\linewidth]{\footnotesize Plain CNN  } \hfill
  \makebox[\wwww\linewidth]{\footnotesize Squeeze\&Excite}\hfill
  \makebox[\wwww\linewidth]{\footnotesize GAMA}\\
  
\subfloat[]{\includegraphics[width=\wwww\linewidth]{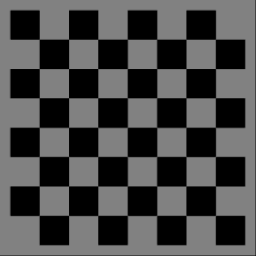}}
\hfill
\subfloat[10.4 dB]{\includegraphics[width=\wwww\linewidth]{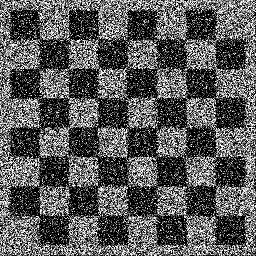}}
\hfill
\subfloat[28.8 dB]{\includegraphics[width=\wwww\linewidth]{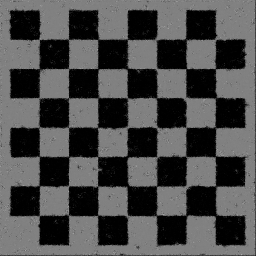}}
\hfill
\subfloat[28.3 dB]{\includegraphics[width=\wwww\linewidth]{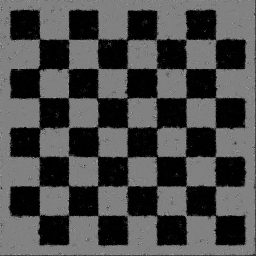}}
\hfill
\subfloat[41.7 dB]{\includegraphics[width=\wwww\linewidth]{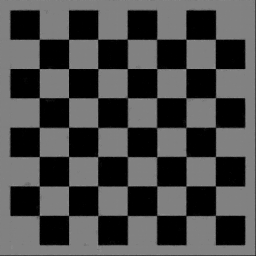}}
\hfill 
\subfloat[]{\includegraphics[width=\wwww\linewidth]{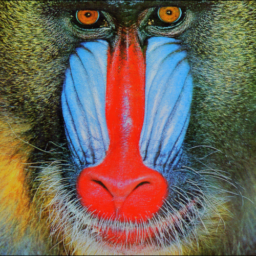}}
\hfill
\subfloat[ 10.5 dB]{\includegraphics[width=\wwww\linewidth]{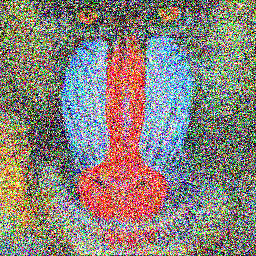}}
\hfill
\subfloat[23.6 dB]{\includegraphics[width=\wwww\linewidth]{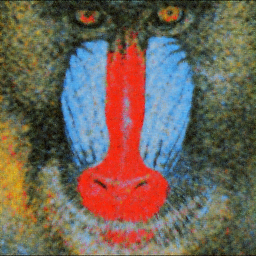}}
\hfill
\subfloat[23.5 dB]{\includegraphics[width=\wwww\linewidth]{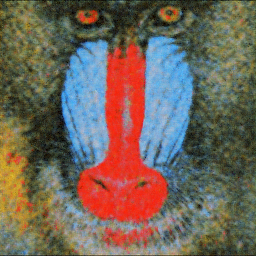}}
\hfill
\subfloat[25.6 dB]{\includegraphics[width=\wwww\linewidth]{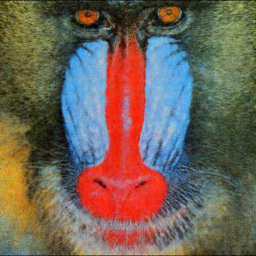}}
\caption{The ability of the underparameterized network illustrated above with different choices for the block to memorize an image with more pixels than it has parameters. The block in the network is either Squeeze\&Excite, or GAMA, or identity in case of the Plain CNN. The total number of parameters is the same for each network variant.
The chess board has a repetitive pattern that the networks with the larger receptive fields can exploit, whereas the baboon has less such structure, which is why we see similar denoising performance. 
}
\label{fig:receptive_field}
\end{figure*}


The receptive field of a convolutional neural network is the spatial extent of the input volume that influences a given pixel in the output volume. 
A large receptive field allows an output pixel to have long range dependencies on far away pixels. 
Deep convolutional networks have a large receptive field \cite{visualize2, understand_cnns}, but are slow since information flows sequentially from layer to layer.

The GAMA block enables a relatively shallow and therefore potentially fast network to have a large receptive field relative to existing blocks. To demonstrate this, we perform the following experiment. 

We consider three variants of a shallow network, as illustrated in Figure \ref{fig:receptive_field}. 
\textbf{Plain CNN}: The plain CNN with only one hidden layer, as displayed in Figure~\ref{fig:receptive_field} with Block=Identity. 
\textbf{Squeeze\&Excite: } The same network, but with the Squeeze-and-Excite block inserted in the hidden layer, i.e., block=Squeeze\&Excite. 
\textbf{GAMA}: The same network again, but block=GAMA block. 

We do not consider the Non-Local blocks in this comparison, since they are slow and memory consuming as seen in the next section with Restormer \cite{restormer}.    

We train each of the three networks to map a single noisy image to its clean counterpart, and measure how well it can recover the clean image. Note that we train and evaluate on the same single image, but since the network is underparameterized (has much less parameters than the image's pixels), it is not able to memorise the exact clean image (achieve 0 training loss). For a fair comparison, the number of channels in the network's hidden layer is adapted so that each of the three networks has the same number of parameters.     

The comparison is shown in figure \ref{fig:receptive_field}. We consider two different images: The first, a chess board, features repeated patterns, while the second, a baboon, lacks such a pattern. A network with a small receptive field operates at a local level, unable to recognize the global patterns present in the chess image. Conversely, a larger receptive field enables the network to perceive the entire chess board as a composition of several identical smaller segments, significantly enhancing pattern recognition and feature extraction. This is evidenced by our GAMA block achieving markedly higher PSNR on the chess image. In contrast, for the Baboon image, where recognisable patterns are absent, the benefits of an enlarged receptive field diminish, resulting in a less pronounced difference in PSNR. The comparative performance on these two images suggests that the improvement observed with the GAMA block on the chess image might be attributed to its larger receptive field.

In this experiment, the blocks were inserted in a small network with one hidden layer, and trained and tested on the same single image to study the receptive field of those blocks. When a deeper network is trained on a large training set and tested on a separate test set, as indented, we also find that a GAMA-IR block works well in a shallow network, but a Squeeze-and-Excite block (as used in NAFNet) does not not work well in a shallow network, as shown in the ablation studies of the original NAFNet paper. 



\section{Experiments}

\newcommand\w{0.195}
\begin{figure*}[t]
  \captionsetup[subfigure]{labelformat=empty,justification=centering}
  \centering
  
\subfloat[Clean]{\includegraphics[width=\w\linewidth]{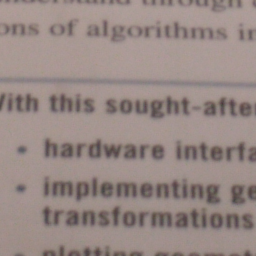}}
\hfill
\subfloat[Noisy: 19.76]{\includegraphics[width=\w\linewidth]{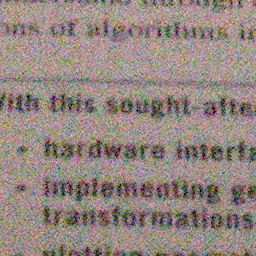}}
\hfill
\subfloat[NAFNet: 36.98 ]{\includegraphics[width=\w\linewidth]{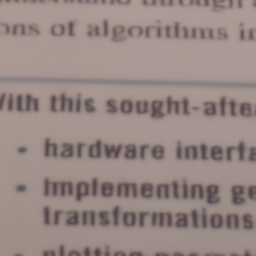}}
\hfill
\subfloat[Restormer: 36.60 ]{\includegraphics[width=\w\linewidth]{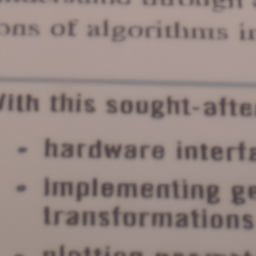}}
\hfill
\subfloat[GAMA-IR: 37.80 ]{\includegraphics[width=\w\linewidth]{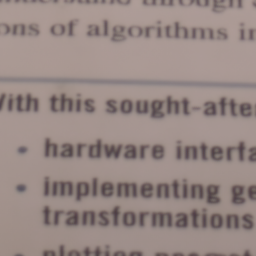}}
\hfill
  \subfloat[Clean]{\includegraphics[width=\w\linewidth]{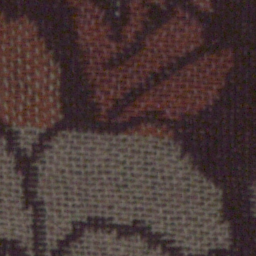}}
  \hfill
\subfloat[Noisy: 17.92]{\includegraphics[width=\w\linewidth]{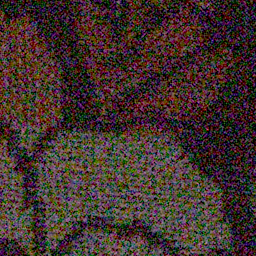}}
\hfill
\subfloat[NAFNet: 32.36]{\includegraphics[width=\w\linewidth]{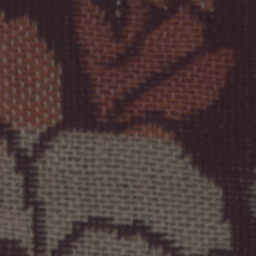}}
\hfill
\subfloat[Restormer: 32.48]{\includegraphics[width=\w\linewidth]{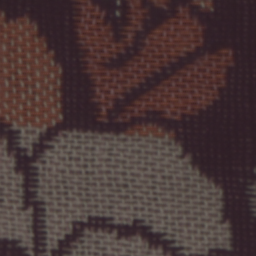}}
\hfill
\subfloat[GAMA-IR: 33.16]{\includegraphics[width=\w\linewidth]{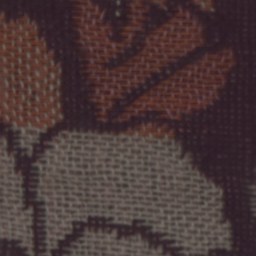}}    
    \caption{ Sample denoised images from SIDD \cite{sidd}. Our method produces sharp images that are almost indistinguishable from the ground truths. 
    }
    \label{fig:sidd}
\end{figure*}

\begin{table}[t]
    \centering
    \resizebox{\columnwidth}{!}{
    \begin{tabular}{|c|c|c|c|c|c|c|c|c|c|c|c|}
    \hline
& \small{MPRNet} & SRMNet & HINet & Uformer & MAXIM & Restormer & NAFNet & GAMA-IR \\ 
& \cite{mprnet} & \cite{srmnet} & \cite{hinet} & \cite{uformer} & \cite{maxim} & \cite{restormer} & 32/64 \cite{nafnet} & S/L (ours) \\ 
\hline
Memory/GB & 6.50 & 3.95 & 2.83 & 4.22 & 13.60 & 13.85 & 2.96/4.52 & 2.98/4.57 \\
Latency/ms & 35.6 & 29.5 & 27.0 & 34.6 & 133.7 & 92.1 & 30.1/35.2 & 15.1/20.7 \\
\hline
PSNR & 39.71 & 39.72 & 39.99 & 39.89 & 39.96 & 40.02 & 39.96/40.30 & 40.02/\bf{40.41} \\
SSIM & 0.958 & 0.959 & 0.958 & 0.960 & 0.960 & 0.960 & 0.960/\bf{0.962} & 0.960/\bf{0.962} \\
\hline
    \end{tabular}} 
    \caption{Denoising results on SIDD \cite{sidd}. Our large model boosts the state-of-the-art PSNR score by 0.11 dB at higher speed. 
    }
    \label{tab:sidd}
\end{table}

We evaluate GAMA-IR on four common image restoration tasks: real-world camera denoising, synthetic Gaussian denoising, motion deblurring, and deraining. For all tasks, we use the  AdamW optimizer \cite{adamw} with initial learning rate $1e^{-3}$ and $\beta_1 = \beta_2 = 0.9$ and a cosine annealing scheduler \cite{cosine_anneal} to reduce the learning rate to $1e^{-7}$. 

We consider a small and large network. 
The small network GAMA-IR-S  has depth 18 and width 42, and the large network GAMA-IR-L has depth 19 and width 80. The small network is trained for 6.4 million iterations and the large network for 25.6 million iterations. 

As performance metrics, we report the PSNR in dB and the SSIM \cite{ssim} scores. As efficiency metrics, we measure the latency and memory on a NVIDIA RTX A6000 GPU at inference for input size $ 1 \times 3 \times 256 \times 256$. Latency and memory are measured as the average of 100 runs after ignoring the first 10 runs, that might fluctuate in speed and memory usage. All networks are run in Pytorch. 

\subsection{Real-world Camera Noise}

We consider the Smartphone Image Denoising Dataset (SIDD) \cite{sidd}, which is the most commonly used dataset for naturally occurring camera noise. It consists of noisy images captured by several smartphone cameras under different lighting conditions and noise patterns. The ground truths are estimated by averaging multiple noisy instances of the same static scene. We take the medium dataset as the training set, and the validation set as the test set. 

The results are in table~\ref{tab:sidd} and figure \ref{fig:sidd}. GAMA-IR improves over the previous state of the art Restormer and NAFNet, while being faster, while being 7 and 2 times faster respectively. 

\newcommand\ww{0.195}
\begin{figure*}[t]
  \captionsetup[subfigure]{labelformat=empty,justification=centering}
  \centering
  
\subfloat[Clean]{\includegraphics[width=\ww\linewidth]{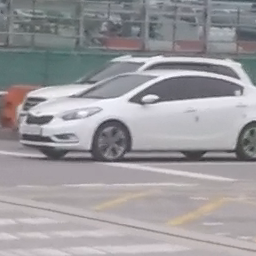}}
\hfill
\subfloat[Blurry: 21.53 dB ]{\includegraphics[width=\ww\linewidth]{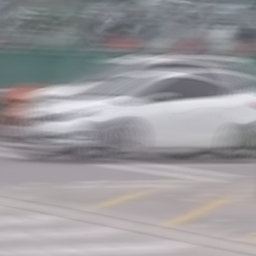}}
\hfill
\subfloat[NAFNet: 26.29 dB ]{\includegraphics[width=\ww\linewidth]{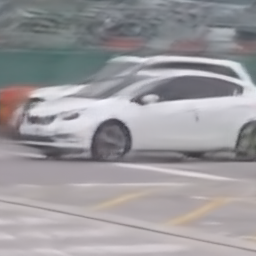}}
\hfill
\subfloat[Restormer: 26.73 dB ]{\includegraphics[width=\ww\linewidth]{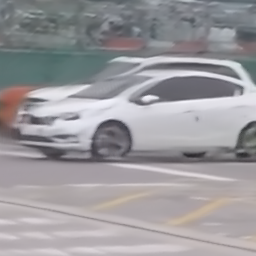}}
\hfill
\subfloat[GAMA-IR: 28.48 dB]{\includegraphics[width=\ww\linewidth]{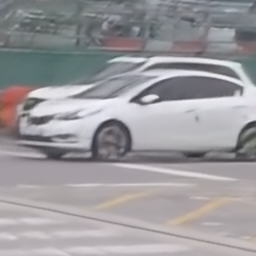}}
\hfill
\subfloat[Clean]{\includegraphics[width=\ww\linewidth]{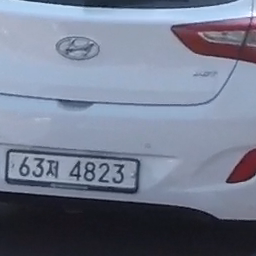}}
\hfill
\subfloat[Blurry: 24.59 dB ]{\includegraphics[width=\ww\linewidth]{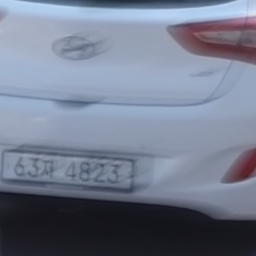}}
\hfill
\subfloat[NAFNet: 35.20 dB ]{\includegraphics[width=\ww\linewidth]{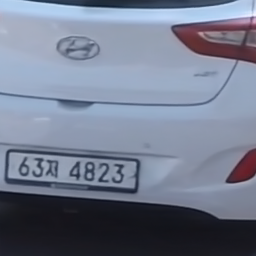}}
\hfill
\subfloat[Restormer: 36.05 dB ]{\includegraphics[width=\ww\linewidth]{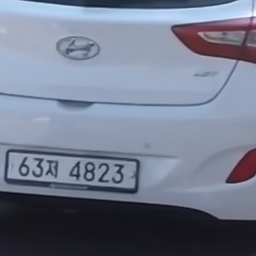}}
\hfill
\subfloat[GAMA-IR: 36.79 dB]{\includegraphics[width=\ww\linewidth]{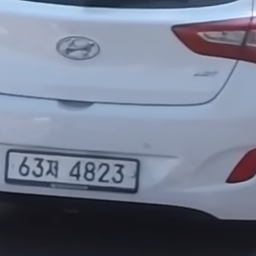}}   
    \caption{Example deblurred results for GoPro \cite{gopro}. Our GAMA-IR network is able to deblur the images well and restores fine details.
    }
    \label{fig:gopro}
\end{figure*}

\begin{table}[t]
    \centering
    \small
    \resizebox{\columnwidth}{!}{
\begin{tabular}{|c|c|c|c|c|c|c|c|c|c|}
\hline
& \small{MIMO-UNet+} & Uformer & MPRNet & HINet & MAXIM & Restormer& NAFNet 32/64 & GAMA-IR \\

 &  \cite{mimo_unet} & \cite{uformer} & \cite{mprnet} & \cite{hinet} & \cite{maxim} & \cite{restormer} & \cite{nafnet}& S/L (ours) \\ 
\hline
Memory/GB & 3.11 & 4.22 & 6.50 & 2.83 & 13.60 & 13.85& 2.96/4.52 & 2.98/4.57 \\
Latency/ms  & 17.0 & 34.6 & 35.6 & 27.0 & 133.7 & 92.1 & 30.1/35.2 & 15.1/20.7 \\
\hline
GoPro  & 32.45 & 32.97 & 32.66 & 32.77 & 32.86 & 32.92 & 32.29/33.08& 32.44/\bf{33.15} \\
\cite{gopro}  & 0.957 & 0.967 & 0.959 & 0.959 & 0.961 & 0.961 & 0.956/\bf{0.963}& 0.958/\bf{0.963} \\
\hline
HIDE & 29.99 & 30.83 & 30.96 & 30.32 & \bf{32.83} & 31.22 & - & 30.57/31.14 \\
\cite{hide}  & 0.930 & 0.952 & 0.939 & 0.932 & \bf{0.956} & 0.942 & -&  0.936/0.943 \\
\hline
\end{tabular}}

    \caption{Deblurring results in terms of PSNR and SSIM for training on GoPro and testing on GoPro and HIDE. 
    }
    \label{tab:deblurring}
\end{table}
\subsection{Deblurring}

We train on the GoPro \cite{gopro} training set and test on the GoPro test set and the HIDE \cite{hide} dataset. The model trained on GoPro is tested on HIDE without any finetuning.  We test on the full resolution test images without any test time refinements such as testing by patches \cite{hinet} or test time local conversion \cite{tlc} for a fair comparison with the competing methods. 


The results are reported in table \ref{tab:deblurring} and figure \ref{fig:gopro}. On GoPro \cite{gopro}, our network achieves best results. On HIDE \cite{hide}, only MAXIM \cite{maxim} outperforms our network, but at the cost of significantly more compute.

\subsection{Deraining}
\newcommand\www{0.24}
\begin{figure}[t]
  \captionsetup[subfigure]{labelformat=empty,justification=centering}
  \centering
  
\subfloat[Clean]{\includegraphics[width=\www\linewidth]{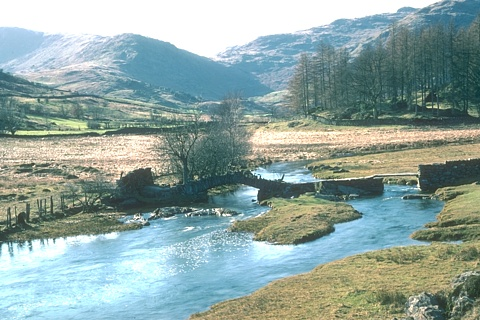}}
\hfill
\subfloat[Noisy: 8.66 dB]{\includegraphics[width=\www\linewidth]{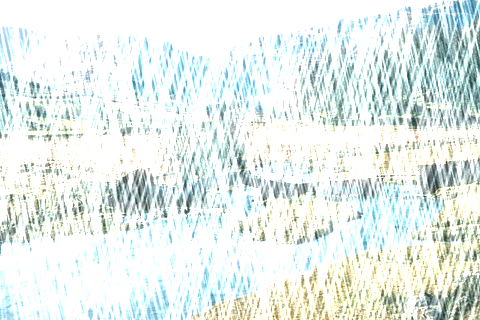}}
\hfill
\subfloat[Restormer: 19.89 dB]{\includegraphics[width=\www\linewidth]{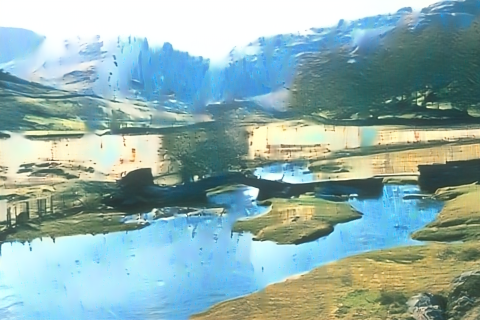}}
\hfill
\subfloat[GAMA-IR: 21.98 dB]{\includegraphics[width=\www\linewidth]{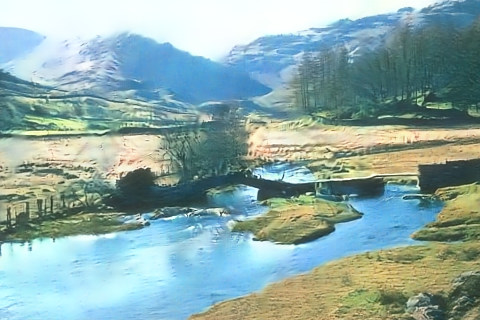}}
  
    \caption{ Derained samples from Rain100H \cite{rain100}. GAMA-IR is able to recover most of the details despite the input image being extremely noisy.
    }
    \label{fig:derain}
\end{figure}
\begin{table}[t]
    \centering
    \small
   \resizebox{\columnwidth}{!}{ 
\begin{tabular}{|c|c|c|c|c|c|c|}
\hline
& SPAIR & MPRNet & HINet & MAXIM & Restormer & GAMA-IR-L \\
& \cite{spair} & \cite{mprnet} & \cite{hinet} & \cite{maxim} & \cite{restormer} & (ours) \\
\hline
\small{Memory/GB} & - & 6.50 & 2.83 & 13.60 & 13.85 & 4.57 \\
\small{Latency/ms} & - & 35.6 & 27.0 & 133.7 & 92.1 & 20.7 \\
\hline
Rain 100H & 30.95 & 30.41 & 30.63 & 30.81 & 31.46 & \bf{33.23} \\
\cite{rain100} & 0.892 & 0.890 & 0.893 & 0.903 & 0.904 & \bf{0.922} \\
\hline
Rain 100L & 36.93 & 36.40 & 37.20 & 38.06 & \bf{38.99} & 37.67 \\
\cite{rain100} & 0.969 & 0.965 & 0.969 & 0.977 & \bf{0.978} & 0.972 \\
\hline
\end{tabular}}
    \caption{PSNR and SSIM for image deraining. Memory and latency measurements for SPAIR are missing, since the code is not available online. 
    }
    \label{tab:deraining}
\end{table}

Following recent works on image deraining \cite{mprnet, maxim, restormer}, we set up a training set collected from several deraining datasets \cite{rain, test100, test1200, test2800}. We evaluate on the two common test sets: Rain100H and Rain100L \cite{rain100}. We calculate the PSNR/SSIM scores by utilizing the Y channel within the YCbCr color space, which is the common evaluation approach in deraining.

As seen in table \ref{tab:deraining} and figure \ref{fig:derain}, our method achieves competitive performance with the state of the art networks, and significantly outperforms them on the Rain100H dataset.

\subsection{Gaussian Noise}

Following \cite{swin_ir, restormer}, we use images from the DIV2K \cite{div2k}, Flickr2K, BSD500 \cite{bsd500}, and WaterlooED \cite{wed} datasets as training images. As test sets, we take the standard benchmarks Kodak24 \cite{kodak24}, and CBSD68 \cite{cbsd68}. 

We consider three noise levels with $\sigma=15, 25, 50$. We show the results in table \ref{tab:gaussian}. On average, GAMA-IR's performance is within state-of-the-art, while being significantly faster and requiring less memory.


As seen from the results on the various tasks, our network's performance is within state-of-the-art, and sometimes even slightly better. Most of the recently proposed state-of-the-art restoration networks achieve minor differences in PSNR/SSIM scores, which are hard to notice when looking at the restored images (see figures \ref{fig:sidd},\ref{fig:gopro},\ref{fig:derain}). Therefore, what makes our network stand out is its low latency, and memory consumption.

\begin{table*}[t]
    \centering
    \resizebox{\columnwidth}{!}{ 
    \begin{tabular}{|@{ }c@{ }|c@{\hspace{0.1cm}}c@{\hspace{0.1cm}}|c@{ }c@{ }c|c@{ }c@{ }c|c@{ }c@{ }c|}
    \hline
    \multirow{2}*{Method}&\small{Memory}&\small{Latency}&\multicolumn{3}{c|}{Kodak24 \cite{kodak24}}&\multicolumn{3}{c|}{CBSD68 \cite{cbsd68}}&\multicolumn{3}{c|}{Average} \\ 
    &/GB&/ms&$\sigma=15$&$\sigma=25$&$\sigma=50$&$\sigma=15$&$\sigma=25$&$\sigma=50$&$\sigma=15$&$\sigma=25$&$\sigma=50$\\
    \hline 
  
 DnCNN\small{\cite{dncnn}}&2.10&5.1&34.60&32.14&28.95&33.90&31.24&27.95&34.25&31.69&28.45 \\
  BRDNet\small{\cite{brdnet}}&2.20&6.2&34.88&32.41&29.22&34.10&31.43&28.16&34.49&31.92&
28.69\\
  SwinIR\small{\cite{swin_ir}}&13.64&149.7&35.34&32.89&29.79&34.42&31.78&28.56&34.88&32.34&29.18 \\  Restormer\small{\cite{restormer}}&13.85&92.1&35.47&33.04&30.01&34.40&31.79&28.60&34.94&\textbf{32.42}&29.31 \\
  GRL-T\cite{GRL}&7.62&72.7&35.24&32.78&29.67&34.30&31.66&28.45&34.77&32.22&29.10 \\
  GRL-B \cite{GRL}&41&503&35.43&33.02&29.93&34.45&31.82&28.62&34.94&\textbf{32.42}&29.28 \\
  GAMA-IR-L&4.57&20.7&35.47&33.04&30.02&34.42&31.79&28.61&\textbf{34.95}&\textbf{32.42}&\textbf{29.32} \\
    \hline
    \end{tabular}}
    \caption{PSNR scores for denoising synthetic Gaussian noise.
    }
    \label{tab:gaussian}
\end{table*}

\section{Ablation Studies}

We perform ablation studies with our small model on SIDD \cite{sidd} denoising and GoPro \cite{gopro} deblurring. We train the model for 6.4 million iterations for every ablation. Results are reported in table \ref{tab:ablation}.

Our GAMA block performs global averaging individually across each dimension, leading to three 2D feature maps (figure \ref{fig:blocks} c). We test with a variant which averages along two dimensions simultaneously as in Squeeze-and-Excite (figure \ref{fig:blocks} b) resulting in three one-dimensional vectors of length $C, H, W$. Each vector is then processed with a separate MLP. This variant is denoted by \textbf{1D Vectors}, and performs worse than the default one, specially on GoPro. It also comes with a signifcant increase in parameter count as discussed in section \ref{sec:global_blocks}.

Another variant we test is a multiplicative instead of an additive operation within the GAMA block. The default version adds all branches together with the input (figure \ref{fig:blocks} c). We replace the addition with a multiplication, and denote this variant with \textbf{Multiplicative}. The default version performs better, is 10\% faster, and consumes 7\% less memory. 

We also test a version of our network without the GAMA block, which we denote by \textbf{w/o GAMA}. 
By comparing its performance to the default version with the GAMA block, we see the importance of increasing the receptive field, and capturing long range dependencies. This variant without the GAMA block is 5\% faster, and consumes 5\% less memory. This shows that the GAMA block can improve performance with minimal computational overhead. 

Additionally, we compare to a UNet \cite{unet}, which is a simple encoder-decoder network consisting of only convolutions, instance normalization \cite{instance_norm}, and ReLU \cite{relu} non-linearities. We create a version of the UNet that consumes the same memory as our default small model, but is twice as fast. As seen from the scores, it performs much worse. 

Finally, we design a faster version of our small model, by decreasing the depth. We increase the width to keep memory consumption constant. This yields a network that is similar to the UNet in both memory, and speed. This variant, denoted  by \textbf{Default (fast)}, outperforms the UNet despite consuming the same compute (memory and speed). 


\begin{table}[h]
    \centering
    \resizebox{\columnwidth}{!}{
    \begin{tabular}{|c|c|c|c|c|c|c|}
    \hline
    &Default&1D Vectors&Multiplicative&w/o GAMA& UNet& Default(fast)\\
    \hline
    SIDD&40.02/0.960&40.00/0.960&39.67/0.958&39.79/0.958&39.43/0.957&39.85/0.959 \\
    GoPro&32.41/0.957&32.26/0.956&31.04/0.943&32.21/0.956&30.28/0.934&31.56/0.949 \\
    \hline
    \end{tabular}}
    \caption{PSNR and SSIM scores of the ablation studies. Most variants have similar latency and memory requirements. Only the UNet and Default(fast) are twice as fast, with same memory consumption.
    }
    \label{tab:ablation}
 
\end{table}


\section{Conclusion}

In this paper we proposed an image restoration network, GAMA-IR, that achieves state-of-the-art restoration performance with low memory consumption and significantly better latency. 
The GAMA block of GAMA-IR mitigates the long latency and high memory consumption associated with deep networks and attention mechanisms used in competing architectures. 
By implementing global additive multidimensional averaging (GAMA) operations, we efficiently capture long range dependencies, and increase the receptive field of a shallow network with minimal computational overhead. The github page for our project is \href{https://github.com/MLI-lab/GAMA-IR}{here}.


\section*{Acknowledgements}

The authors are supported by the German Federal Ministry of Education and Research and the Bavarian State Ministry for Science and the Arts. The authors of this work take full responsibility for its content.
The authors also received funding by the Deutsche Forschungsgemeinschaft (DFG, German
Research Foundation) - 456465471, 464123524.

{ 
\AtNextBibliography{\small} 
\printbibliography
}

\end{document}